\documentclass[
]{ceurart}

\usepackage{listings}
\usepackage{xcolor}
\usepackage{tikz}
\makeatletter
\newcommand{\displaybump}{\hbox to \@totalleftmargin{\hfil}}
\makeatother
\usepackage{amsmath}
\DeclareMathOperator*{\argmin}{arg\,min}
\DeclareMathOperator*{\argmax}{arg\,max}
\begin{document}

\copyrightyear{2024}
\copyrightclause{Copyright for this paper by its authors.
  Use permitted under Creative Commons License Attribution 4.0
  International (CC BY 4.0).}

\conference{CHR 2024: Computational Humanities Research Conference, December 4–6, 2024, Aarhus, Denmark}

\title{Context is Key(NMF): Modelling Topical Information Dynamics in Chinese Diaspora Media}

\author[1,2]{Ross Deans Kristensen-McLachlan}[%
orcid=0000-0001-8714-1911,
email=rdkm@cc.au.dk,
]
\cormark[1]
\author[1,3]{Rebecca M. M. Hicke}[%
orcid=0009-0006-2074-8376,
email=rmh327@cornell.edu,
url=https://rmatouschekh.github.io,
]
\author[1]{Márton Kardos}[%
orcid=0000-0001-9652-4498,
email=martonkardos@cas.au.dk,
]
\author[4]{Mette Thunø}[%
orcid=0000-0001-9410-2363,
email=mettethunoe@cas.au.dk,
]
\address[1]{Center for Humanities Computing, Aarhus University, Denmark}
\address[2]{Department of Linguistics, Cognitive Science, and Semiotics, Aarhus University, Denmark}
\address[3]{Department of Computer Science, Cornell University, USA}
\address[4]{Department of Global Studies, Aarhus University, Denmark}

\cortext[1]{Corresponding author.}

\begin{abstract}
Does the People's Republic of China (PRC) interfere with European elections through ethnic Chinese diaspora media? This question forms the basis of an ongoing research project exploring how PRC narratives about European elections are represented in Chinese diaspora media, and thus the objectives of PRC news media manipulation. In order to study diaspora media efficiently and at scale, it is necessary to use techniques derived from quantitative text analysis, such as topic modelling. In this paper, we present a pipeline for studying information dynamics in Chinese media. Firstly, we present KeyNMF, a new approach to static and dynamic topic modelling using transformer-based contextual embedding models. We provide benchmark evaluations to demonstrate that our approach is competitive on a number of Chinese datasets and metrics. Secondly, we integrate KeyNMF with existing methods for describing information dynamics in complex systems. We apply this pipeline to data from five news sites, focusing on the period of time leading up to the 2024 European parliamentary elections. Our methods and results demonstrate the effectiveness of KeyNMF for studying information dynamics in Chinese media and lay groundwork for further work addressing the broader research questions.
\end{abstract}


\maketitle

\section{Introduction}

A number of major elections took place in the West over the course of 2024. Across Europe, citizens took to the polls in early June to elect members to the European Parliament. In France, the election of a new \textit{Assemblée nationale} caused political turmoil, while the United Kingdom voted in a Labour government for the first time in 14 years. On the other side of the Atlantic Ocean, the United States of America will vote to determine their new President in November. The fallout of these elections remains to be seen but it seems clear that this year is one of political change and upheaval.

Much digital ink is spilled on these topics in Western media as the various electorates determine their preferences before elections and digest the fallout afterwards. Moreover, a significant part of this media coverage is fundamentally persuasive, aiming to convince voters to bet on the candidate who most closely aligns with the social and economic ideology of the media outlets and their owners \cite{gatterman_2022}. Likewise, coverage of these elections is not limited to European media institutions, with media outlets around the world updating their readership on how these elections impact them. 

In this context, one particular type of media stands out as especially interesting: ethnic Chinese media targeting diaspora communities in Europe, a group which by some estimates comprises around 1.5-3 million individuals. These media outlets are potentially invaluable sources for understanding how the Chinese government and the Chinese Communist Party (CCP) attempt to influence the diaspora. Furthermore, studying these outlets potentially provides unique insights into how China views itself in relation to the West by showing how the PRC presents itself to its diaspora groups. A growing body of literature has already begun to address these questions in the context of social media \cite{schliebs2021, thuno2024initial} or in terms of digital infrastructure more generally \cite{brussee_2022,chan_alden_2023}. In ongoing research, our aim is to assess whether Chinese diaspora news sources intend to impact opinions on elections in the West during 2024. We attempt to understand the control of information flow in Chinese diaspora media and how this control is used to set specific agendas during electoral periods: promoting certain political parties or individual candidates, polarizing citizens, and attacking or promoting specific political positions. 

To pursue this research, we design a pipeline for analyzing large amounts of Chinese-language news data. First, we introduce KeyNMF, a novel approach to creating context-sensitive topics models via transformer-based encoder models. KeyNMF can be trivially applied across different languages and in data scarce environments, and is shown here to create coherent, human-interpretable outputs when working with Chinese language data. We then integrate KeyNMF with existing techniques for describing the information dynamics of complex systems which measure the novelty and resonance of information present in a system over time. We use this pipeline to perform preliminary analysis on our dataset of Chinese diaspora media, finding clear trends in the novelty and resonance signals which correlate with significant political events. The results presented are thus intended to be both a proof of concept and a stepping stone towards more meaningful understanding of the dynamics underlying Chinese diaspora media.

\section{Related Work}

\subsection{Information Dynamics}

The study of information dynamics in complex cultural systems has been a central aspect of research in computational humanities and cultural analytics in recent years. One of the most promising approaches to this problem was introduced in \cite{barron-2018-individuals} which studied the shifting debates which took place during the French Revolution. In this approach, divergence in content between different time slices can be calculated using information-theoretic measures. These measures can then be used to quantify two interrelated values: the \textit{novelty} of the system, or how much the new time slice diverges from preceding time slices; and the \textit{resonance} of this information, which describes how information persists over time. 

Novelty-resonance patterns have been studied in a number of different discourse domains. \cite{nielbo2020trend} demonstrate their usefulness in identifying so-called trend reservoirs on Reddit. Similar interaction patterns between novelty and resonance have been successfully employed to study the manner in which online news media responded to catastrophic events \cite{nielbo2021news, nielbo2021no, nielbo2023pandemic}. In \cite{wevers2021event}, the same fundamental method of analysis demonstrates that novelty-resonance patterns clearly track major social and historical events in the 20th century, using data taken from the front page of Dutch newspapers.

Calculating these underlying dynamics requires the creation of some kind of numerical representation of the data. Specifically, the difference between individual windows is computed by finding the windowed relative entropy, in this case calculated using Jensen-Shannon Divergence (JSD). Since JSD computes the distance between probability distributions, the numerical representations of the data are required to take that form. In \cite{baglini2022emodynamics}, this was achieved by calculating the probabilities of a pre-trained, BERT-based emotion classification model, where the predicted probabilities for each label created a distribution over emotions for each document. However, for most purposes, novelty and resonance are calculated based on distributions generated by a probabilistic topic model. 

\subsection{Vanilla LDA}

Typically, novelty and resonance are calculated from topic probability distributions extracted by Latent Dirichlet Allocation (LDA) \cite{blei_lda,blei_prob_topic_models}.
Topic distributions in documents are a natural choice for information dynamics, as they are immediately usable with entropy-based measures.
LDA is a generative bag-of-words model, which assumes that a document contains a mixture of topics and all words in the document are drawn from this mixture distribution.

However, LDA has a number of well-known shortcomings.
Documents have to be heavily pre-processed for optimal results;
otherwise, the topic descriptions produced by the model are often contaminated by noise and stop words \cite{kardos2024s3semanticsignal}.
In addition, since LDA makes the bag-of-words assumption, it cannot utilize contextual and syntactic information, nor general properties of natural language learned from outside sources.
Finally, LDA is sensitive to hyperparameter choices and \citet{Wallach2009RethinkingLW} demonstrate that using symmetric Dirichlet priors, which is the case in canonical implementations \cite{gensim, scikit-learn} and the majority of academic studies, can lead to sub-optimal performance.

There have also been challenges to the generalizability of LDA from the perspective of Chinese NLP, as the primary structural and semantic unit of Chinese is the character rather than the word \cite{zhao_2011, QIN201660}. While these concerns might be overstated, working with Chinese language data causes specific challenges in terms of tokenization and semantics which directly impact the efficacy of traditional LDA approaches to topic modelling.

\subsection{Alternatives to LDA}

A major shortcoming of LDA when trying to model change over time is that topics are calculated over all documents, essentially flattening any temporal aspect of the data. This is undesirable, since topics themselves naturally evolve over time, meaning that LDA may not reflect the true dynamics of a system. These issue is partly rectified by dynamic topic models \cite{blei_dynamic} which account for temporal changes in topics with a state-space model. However, Dynamic LDA models are even more parameter-rich than the vanilla implementation and thus amplify its limitations.

Recently, contemporary topic models have shown that it is possible to utilize embeddings from the sentence transformers \cite{reimers-2019-sentence-bert} to infuse contextual information into topic models and to allow for transfer learning \cite{ctm, zeroshot_tm, bertopic_paper, top2vec,kardos2024s3semanticsignal}. This contextual information can lead to more coherent and semantically interpretable topics. In addition, since these models draw on existing pre-trained language models, they do not require training a generative model from scratch. This means that it is possible to train topic models in data scarce contexts where traditional LDA might perform poorly.

Among the most popular of these contemporary models is BERTopic \cite{bertopic_paper}, which also has dynamic modelling capabilities. In this model, topic-term importances are estimated post-hoc on pre-defined time slices based on one underlying topic model. However, as with LDA, BERTopic is sensitive to pre-processing \cite{kardos2024s3semanticsignal}. Additionally, because BERTopic is a clustering topic model, documents are only assigned a single topic label. This renders the model impractical in settings where documents are expected to contain multiple topics and means that BERTopic is not suitable for calculating novelty and resonance, since the entropy calculations assume probability distributions over documents. 




\section{KeyNMF}

We propose KeyNMF, a novel topic modelling approach that utilizes neural text embeddings.
KeyNMF builds on the reliability, stability \cite{nmf_stability}, scalability \cite{online_nmf}, and interpretability of Non-negative Matrix Factorization (NMF) \cite{nmf_impl}, while mitigating its sensitivity to pre-processing and making use of contextual information in texts.
This is achieved by: 1) computing keyword importances from documents with contextual embeddings (similar to KeyBERT \cite{grootendorst2020keybert}); and 2) decomposing those importances with NMF.

We release an implementation of KeyNMF as part of the \texttt{Turftopic} Python package.\footnote{\url{https://x-tabdeveloping.github.io/turftopic/}}

\subsection{Model Description}
KeyNMF operationalizes topic extraction as the following steps:
\begin{enumerate}
    \item For each document $d$:
    \begin{enumerate}
        \item Let $x_d$ be the document's embedding produced with an encoder model.
        \item Let $v_w$ be the word embedding of a word $w$ produced with the same encoder model.
        \item Let $K_d$ be the set of $N$ keywords in $d$ with the highest cosine similarity to $d$:
        \[
        K_d = \argmax_{K^*} \sum_{w \in K^*} \text{sim}(x_d,v_w) \text{, where }
        |K_d| = N\text{ and }
        w \in d \displaybump
        \]
    \end{enumerate}
    \item Arrange the keyword similarities into a non-negative keyword matrix $M$.
    Let $M_{dw}$ be the importance of keyword $w$ in document $d$:
    \[
    M_{dw} = 
    \begin{cases}
    \text{sim}(d,w), & \text{if } w \in K_d \text{ and } \text{sim}(x_d,v_w) > 0 \\
    0, & \text{otherwise}.
    
    \end{cases}\displaybump
    \]
    \item Decompose $M$ with non-negative matrix factorization: $M \approx WH$,
    where $W$ is the document-topic matrix, and $H$ is the topic-term-matrix.
    This is achieved with coordinate-descent, minimizing the square loss $L(W,H) = ||X - WH||^2$.
\end{enumerate}

\subsection{Dynamic KeyNMF}
KeyNMF can be used for modelling topics' evolution in a corpus over time.
This is done by first computing a global model over the entire corpus, then calculating time-specific topic-term importances in predefined time slices. Specifically:
\begin{enumerate}
    \item Compute the keyword matrix $M$ for the whole corpus.
    \item Decompose $M$ with non-negative matrix factorization: $M \approx WH$.
    \item For each time slice $t$:
    \begin{enumerate}
        \item Let $W_t$ be a subset of $W$ and $M_t$ a subset of $M$ for the documents in time slice $t$.
        \item Obtain the topic-term-matrix for $t$ with NMF while fixing $W_t$:
        \[
        H_t = \argmin_{H^{*}} ||M_t - W_t H^{*}||^2\displaybump
        \]
        \item The temporal importance of topic $j$ is then $I_{tj} = \sum_{d \in t} (W_{t})_{dj}$, where all $d$ are documents in time slice $t$.
        We can obtain pseudo-topic distributions in the time-slices by L1-normalizing the temporal importances:
        $\hat{P}_{tj} = \frac{I_{tj}}{\sum_{i} I_{ti}}$.
    \end{enumerate}
\end{enumerate}

Since NMF is not a probabilistic model, we use temporal pseudo-probabilities as a proxy for topic distributions.

\subsection{Performance}
To demonstrate KeyNMF's effectiveness as a topic model, we evaluate its performance using the \texttt{topic-benchmark} Python package and the \texttt{paraphrase-multilingual-MiniLM-L12-v2}\footnote{\url{https://huggingface.co/sentence-transformers/paraphrase-multilingual-MiniLM-L12-v2}} embedding model. 15 keywords are extracted for each document.
Our evaluation procedure is based on  that of \citet{kardos2024s3semanticsignal}, but, since our intended use case is Chinese news data, we ran the benchmark using the same corpora and pipeline as in our investigations (see Sections \ref{sec:data} and \ref{sec:methods}).
Additionally, we utilized paraphrase-multilingual-MiniLM for measuring external word embedding coherence, instead of an English Word2Vec model. 
\footnote{This gives Top2Vec an unfair advantage on this metric as it selects descriptive words based on the same criteria as the metric. $C_{ex}$ scores on Top2Vec should thus be interpreted with caution.}

\begin{table}
\caption{KeyNMF's performance on Chinese news data against a number of baselines.
Topic descriptions were evaluated on diversity ($d$), internal ($C_{in}$) and external ($C_{ex}$) word embedding coherence.
}
\resizebox{\textwidth}{!}{
\setlength{\tabcolsep}{4pt}
\begin{tabular}{lccccc}
\toprule
&\textbf{chinanews} & \textbf{ihuawen} & \textbf{oushinet} & \textbf{xinozhou} & \textbf{yidali-huarenjie}\\

\begin{tabular}{l}
\textbf{Model} \\
\midrule
\textbf{KeyNMF} \\
\textbf{S³} \\
\textbf{Top2Vec} \\
\textbf{BERTopic} \\
\textbf{CTM\textsubscript{combined}} \\
\textbf{CTM\textsubscript{zeroshot}} \\
\textbf{NMF} \\
\textbf{LDA} \\
\end{tabular} &
\begin{tabular}{ccc}
$d$ & $C_{\text{in}}$ & $C_{\text{ex}}$\\
\midrule
0.93 & \underline{0.29} & \underline{0.63}\\
0.91 & 0.16 & 0.47\\
0.78 & 0.14 & \textbf{0.71}\\
0.89 & \textbf{0.31} & 0.52\\
\textbf{0.99} & 0.27 & 0.52\\
\underline{0.99} & 0.28 & 0.53\\
0.74 & 0.27 & 0.57\\
0.61 & 0.19 & 0.57\\
\end{tabular} &

\begin{tabular}{ccc}
$d$ & $C_{\text{in}}$ & $C_{\text{ex}}$\\
\midrule
0.91 & 0.17 & \underline{0.64}\\
0.91 & 0.11 & 0.47\\
0.83 & 0.10 & \textbf{0.70}\\
0.89 & \textbf{0.26} & 0.50\\
\textbf{0.99} & \underline{0.23} & 0.51\\
\underline{0.99} & 0.23 & 0.50\\
0.60 & 0.18 & 0.53\\
0.53 & 0.16 & 0.54\\
\end{tabular} &

\begin{tabular}{ccc}
$d$ & $C_{\text{in}}$ & $C_{\text{ex}}$\\
\midrule
0.84 & \textbf{0.23} & \underline{0.58}\\
0.83 & 0.12 & 0.54\\
0.87 & 0.12 & \textbf{0.73}\\
0.84 & \underline{0.23} & 0.50\\
\underline{0.99} & 0.21 & 0.51\\
\textbf{0.99} & 0.22 & 0.50\\
0.64 & 0.18 & 0.54\\
0.41 & 0.13 & 0.54\\
\end{tabular} &

\begin{tabular}{ccc}
$d$ & $C_{\text{in}}$ & $C_{\text{ex}}$\\
\midrule
0.85 & \underline{0.26} & 0.55\\
0.96 & 0.17 & 0.55\\
0.86 & 0.14 & \textbf{0.71}\\
0.84 & \textbf{0.26} & 0.52\\
\underline{0.98} & 0.25 & 0.51\\
\textbf{1.00} & 0.26 & 0.51\\
0.66 & 0.18 & 0.56\\
0.48 & 0.14 & \underline{0.58}\\
\end{tabular} &

\begin{tabular}{ccc}
$d$ & $C_{\text{in}}$ & $C_{\text{ex}}$\\
\midrule
0.88 & 0.52 & \underline{0.57}\\
0.93 & 0.46 & 0.52\\
0.75 & 0.46 & \textbf{0.69}\\
0.91 & \textbf{0.57} & 0.51\\
\textbf{0.97} & \underline{0.54} & 0.49\\
\underline{0.97} & 0.54 & 0.51\\
0.71 & 0.49 & 0.54\\
0.57 & 0.34 & 0.54\\
\end{tabular} \\

\bottomrule
\end{tabular}
}
\label{tab:performance}
\end{table}

\begin{figure}[htbp]
    \centering
    \includegraphics[width=1\linewidth]{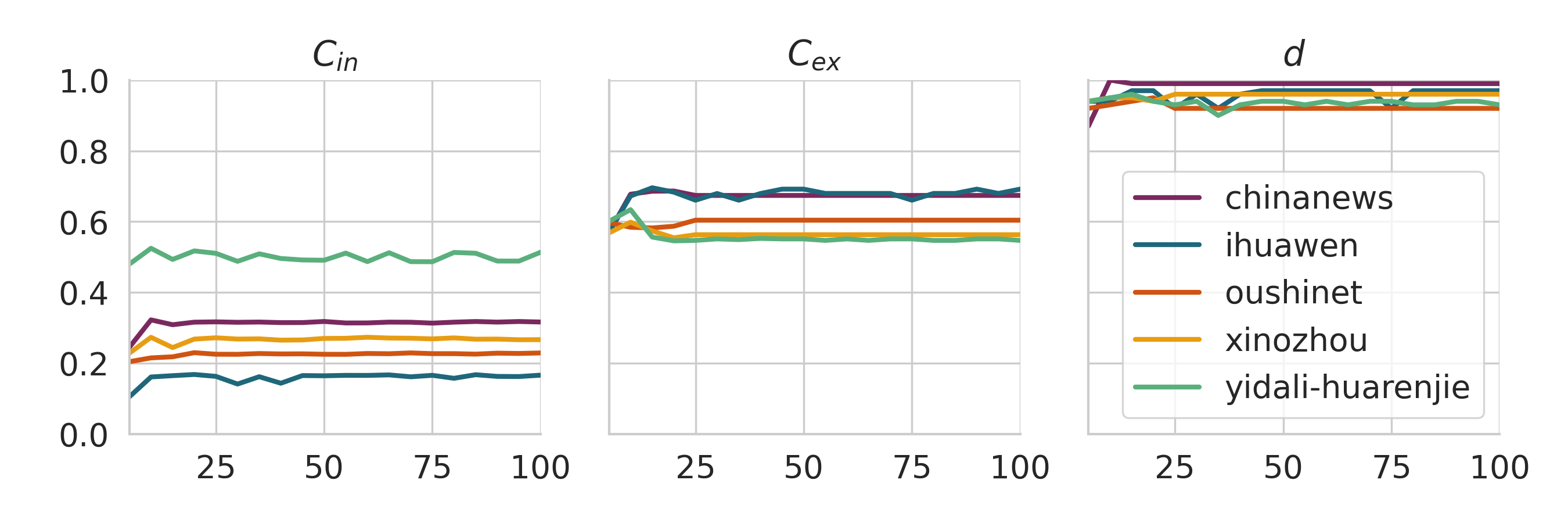}
    \caption{Sensitivity of KeyNMF to the choice of $N$ keywords on multiple metrics and news sources.}
    \label{fig:sensitivity}
\end{figure}

Based on our evaluations, KeyNMF's performance is comparable with state-of-the-art contextual topic models,
and performs especially well on external coherence, only rivalled by Top2Vec on most corpora, which explicitly selects words based on their proximity in semantic space (see Table \ref{tab:performance}).
The model represents a drastic improvement over classical topic models outperforming both NMF and LDA significantly, indicating that the contextual information infused into the model enhances its performance in a meaningful way.

\subsubsection{Sensitivity to Number of Keywords}

We additionally test whether the number of keywords extracted from a text influences the model's performance on different corpora, which allows us to determine KeyNMF's robustness to hyperparameter choices.
We used the same news sources, pipeline, and quantitative metrics for evaluating this property of the model as for previous evaluations and analyses.
The number of keywords was varied from 5 to 100 with a step size of 5 (see Figure \ref{fig:sensitivity}).

We observed that performance was relatively stable regardless of number of keywords, and converged rather quickly.
Only minimal fluctuations are observable with $N > 25$ on most corpora.
However, on Xinozhou and Yidali-Huarenjie, lower values of $N$ (5-15) resulted in higher coherence scores.
We thus deem 15 keywords a balanced choice of $N$ for further investigations.

\section{Data}
\label{sec:data}

Having demonstrated the effectiveness of KeyNMF, we use it as the basis for our study of Chinese diaspora media. Our dataset comprises news articles from five sites aimed at Chinese diaspora populations in the EU: Chinanews,\footnote{\url{http://www.chinanews.se}} Ihuawen,\footnote{\url{https://ihuawen.com}} Yidali Huarenjie,\footnote{\url{https://yidali.huarenjie.com}} Xinouzhou,\footnote{\url{https://www.xinouzhou.com}} and Oushinet.\footnote{\url{http://www.oushinet.com}} We select these sites because they represent a variety of formats, audiences, and perspectives. Oushinet has the largest target audience, with articles in several languages and local journalists writing specifically for the site. In contrast, Xinouzhou reports mostly on Chinese local news, Yidali Huarenjie and Chinanews are community media platforms based in Italy and Scandinavia respectively, and Ihuawen is a weekly magazine based in the United Kingdom.

\begin{figure*}
    \centering
    \includegraphics[height=0.4\linewidth]{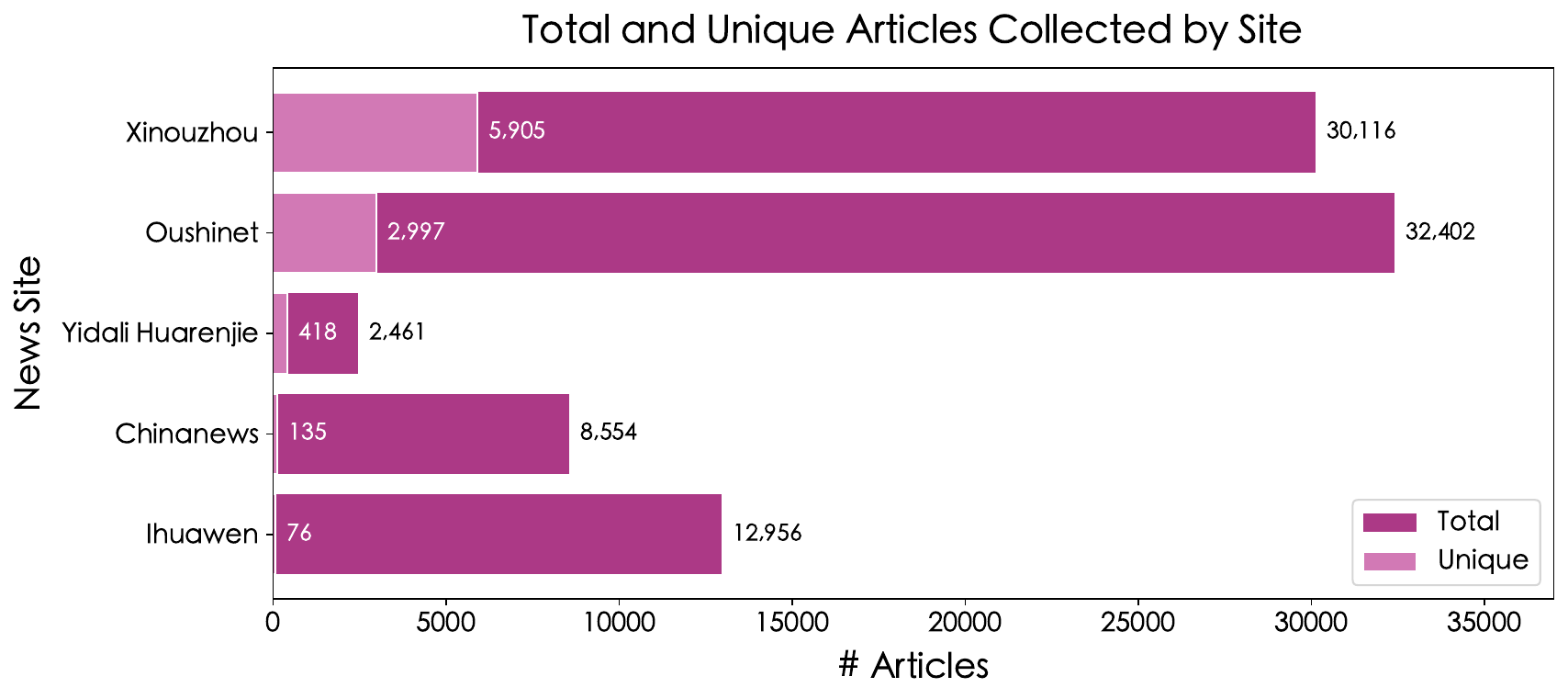}
    \label{fig:datasetSize}
    \caption{The total and unique number of articles collected for each news site.}
    \label{fig:datasetInfo}
\end{figure*}

Our data collection focuses on articles linked from each site's front page and a selection of subpages we deem likely to contain information on international relations, particularly with Europe (listed in Appendix \ref{sec:newsSubpages}). We hypothesize that articles linked from these main pages will reflect the topics each news site is attempting to highlight, and will thus provide information on the priorities of the forces backing the media landscape. We scrape all articles linked from each front page and subpage every six hours using a custom web scraper.
An article is only scraped once per time point, even if it was linked from multiple pages, but can be scraped multiple times if it appears at multiple time points. Data collection from four sites --- Chinanews, Ihuawen, Xinouzhou, and Oushinet --- began at 18:15 on April 30, 2024 and collection from the fifth site, Yidali Huarenjie, began at 12:15 on May 7, 2024. Our dataset includes all articles scraped until 6:15 on June 17, 2024, one week after the EU Parliamentary elections took place. Once scraped, we extract the body of each article from the corresponding html file. We attempt to minimize the amount of boilerplate text (e.g.\ bylines and publication dates) included in the extracted texts; although it is impossible to remove all such text from our dataset, a hand analysis of ten random articles from each news site indicates that the amount of `junk' text included in the final dataset is minimal. 

\begin{figure}
    \centering
    \includegraphics[scale=0.35]{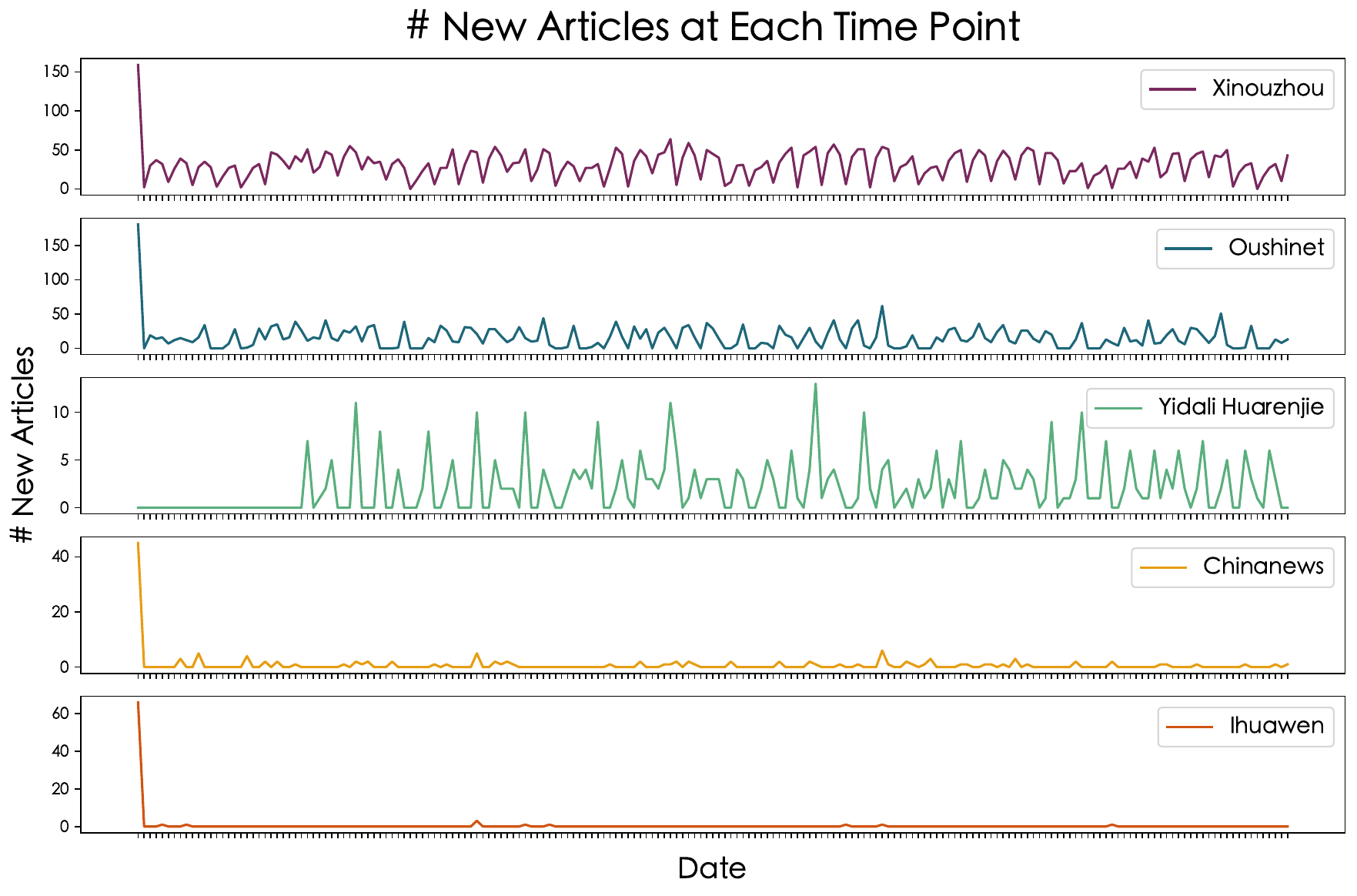}
    \caption{The number of new articles collected at each time point for each source. An article is `new' if it did not appear in the collected set of articles from the previous time point.}
    \label{fig:articleTurnover}
\end{figure}

The total and unique number of articles collected from each site are reported in Figure \ref{fig:datasetInfo}. It is clear that different sites follow different publication patterns. To further validate this, we examine the number of `new' articles at each time point for each source, or the number of articles that were not included in the last scrape (Figure \ref{fig:articleTurnover}). We see that some sites, like Xinouzhou and Yidali Huarenjie, frequently refresh the articles displayed on their main pages, leading to a larger number of unique articles. In contrast, sites like Ihuawen appear to keep several articles on the main pages for a long time, meaning that they display a very small number of unique articles overall. These differences likely affect the patterns we see in the information systems for each source.

\section{Experimental Design}
\label{sec:methods}

Extracted article texts are embedded with a multilingual transformer-based model \cite{reimers-2020-multilingual-sentence-bert}\footnote{\texttt{paraphrase-multilingual-MiniLM-L12-v2}} using the Sentence Transformers library.\footnote{\url{https://sbert.net}} The embedding is done entirely on a 64-core CPU with 384GB RAM. Each document is embedded once for each time it appears in the dataset. In total, embedding all the documents takes $\sim$2 hours. The maximum sequence length of this embedding model is 128 tokens. Thus, any article longer than 128 tokens is truncated and information from later in the piece is not included in the embedding. Although this is a limitation, we do not consider it prohibitive, as previous research has shown that the bulk of the content in a news article is presented at the very beginning --- a widely-practiced professional standard for journalistic writing known as the \textit{inverted pyramid} \cite{pottker_2003}. 

Since our primary interest is understanding the evolution of information dynamics in each news site over time, we use Dynamic KeyNMF to find topic proportions for each timeslice. For keyword extraction, we utilize the \texttt{jieba} tokenizer and remove stop-words present in an authoritative list,\footnote{\url{https://github.com/stopwords-iso/stopwords-zh/blob/master/stopwords-zh.txt}} with the retained tokens then encoded using the same multilingual model as was used on the documents \cite{reimers-2020-multilingual-sentence-bert}. We fit multiple models with 10, 25, and 50 topics respectively in order to investigate topical dynamics at multiple levels of granularity. Separate models are fit for each news site. The plotted topics over time, top keywords for each topic at each timeslice, and topic distributions at each timeslice are extracted from each model and saved for further analysis.

We then use the topic pseudo-distributions to measure the novelty and resonance signals for each news site and, following \cite{nielbo2021no} and \cite{baglini2022emodynamics}, use windowed relative entropy with Jensen-Shannon divergence to calculate both metrics.
For a window of size $n$, the novelty at time point $t$ is the mean entropy of the topic pseudo-distribution at $t$ ($\hat{P}_t$) and the $n$ previous pseudo-distributions.
The transience at time point $t$ is the mean entropy of the topic pseudo-distribution at $t$ and the $n$ subsequent pseudo-distributions.
Then, the resonance of a time point is the novelty at that point minus the transience.
We use a window of size 12 when calculating both signals, which is equivalent to three days of data.

We apply nonlinear adaptive filtering to smooth the extracted novelty and resonance, again following \cite{nielbo2021no} and \cite{baglini2022emodynamics}.
This removes noise from the signals by calculating the value at a given time point relative to the surrounding time points. We use a span of 56, the same as \cite{baglini2022emodynamics}, for smoothing. The code we use for calculating novelty and resonance is adapted from that released alongside \cite{baglini2022emodynamics} and \cite{nielbo2021no}.

\section{Results and Discussion}

\begin{figure}
\centering
\includegraphics[scale=0.19]{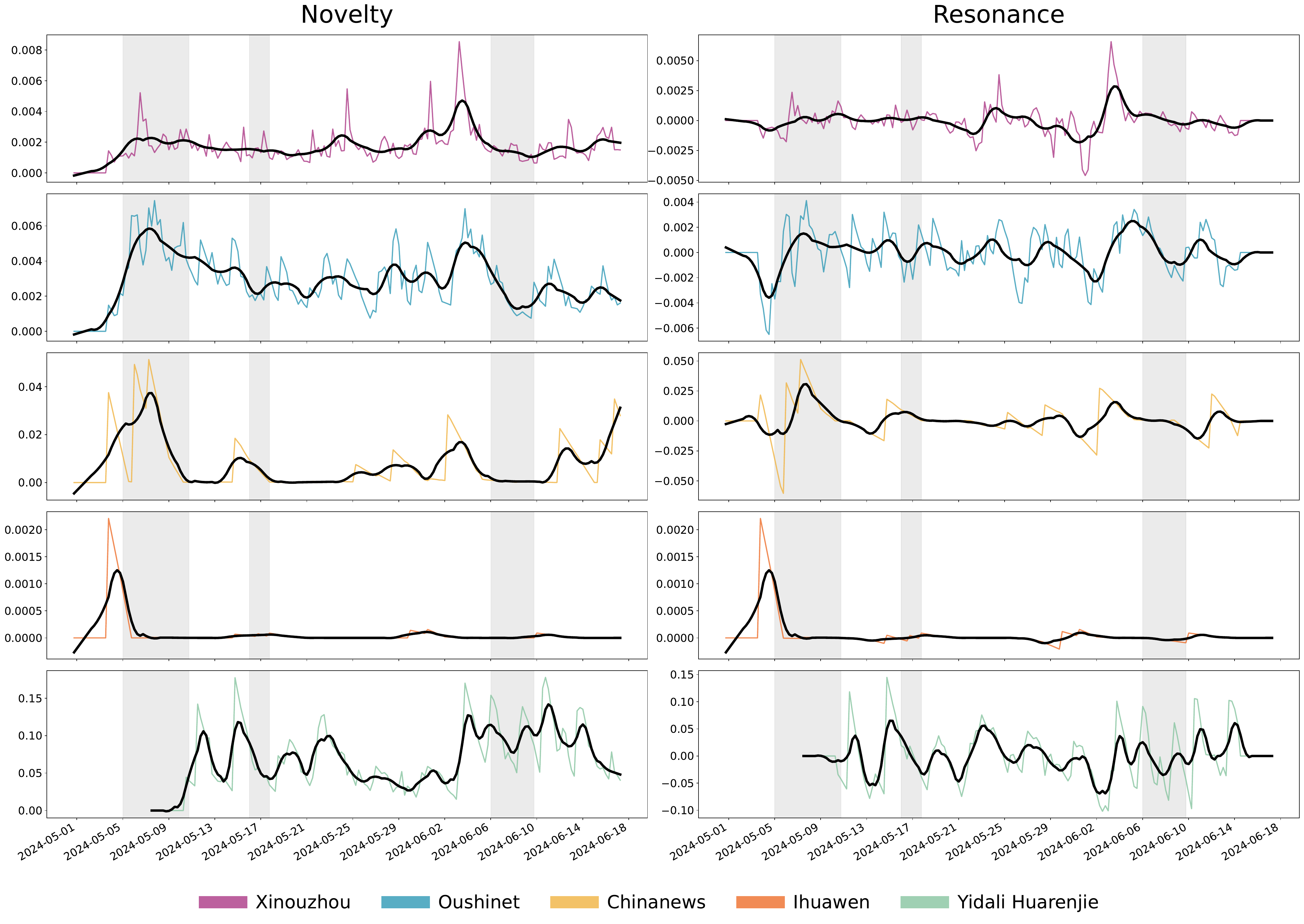}
\caption{The novelty and resonance plots for each news site from KeyNMF with ten topics. The three shaded areas represent Xi Jinping's European tour (May 5-10, 2024), Putin's state visit to China (May 16-17, 2024), and the EU parliamentary elections (June 6-9, 2024). Note that the y-axis ranges differ for each chart.}
\label{fig:nov-res-10}
\end{figure}

We find clear trends in the novelty and resonance signals that correlate to significant events in the EU during the period studied: Xi Jinping's European Tour (May 5-10), Putin's state visit to China (May 16-17), and the EU parliamentary elections (June 6-9). Our analysis focuses on the novelty and resonance trends extracted from the KeyNMF models with ten topics as these provide the clearest signals. The results for 25 and 50 topics are included in Appendix \ref{sec:novelty-resonance-ablations}. We additionally focus our in depth discussion of the results on the two largest news sources, Xinouzhou and Oushinet, for this preliminary validation of the pipeline.

We see spikes in novelty of varying strengths for both Xinouzhou and Oushinet during Xi Jinping's European tour (Figure \ref{fig:nov-res-10}). There are also corresponding dips in resonance before his tour for both sites, followed by increases in resonance during the tour. This indicates that novel information is introduced to the site ecosystems during the tour which replaces previous topics of interest, and which persists in the system for some time.

One of the most productive aspects of Dynamic KeyNMF is that it allows us to study topic fluctuations over time. Thus, we explore which topical shifts contribute to changes in the novelty and resonance signals. For example, on Oushinet, the time period during Xi Jinping's European tour is associated with high pseudo-probabilities for a topic defined by the keywords \textit{Paris, France} and \textit{state visit} and a topic defined by \textit{President}, \textit{China}, and \textit{Xi Jinping} (Appendix \ref{sec:topic-distributions}, Figure \ref{fig:tour-topics}). Towards the end of the tour, a topic on diplomacy and \textit{bilaterial relations} between China and France also gains prominence. For Xinouzhou, this time period contains a peak in the pseudo-probabilities for two topics on Hungary and Chinese relations with Hungary, one of the locations on the tour.

Similarly, there is a noticeable spike in the novelty and resonance for Oushinet directly before Putin's state visit to China. This period is marked by relatively high pseudo-probabilities for a topic characterized by the terms \textit{China}, \textit{Beijing}, \textit{Chinese}, and \textit{Chinese News Service} and a topic with the keywords \textit{Russia}, \textit{Ukraine}, \textit{Putin}, and \textit{Moscow} (Appendix \ref{sec:topic-distributions}, Figure \ref{fig:state-visit-topics}).

Most significantly for this study, there are fluctuations in novelty and resonance for both sites around the EU parliamentary elections. Specifically, there are peaks in the novelty and resonance signals for Xinouzhou and Oushinet before and after the elections, with troughs throughout much of the election period. We hypothesize that these trends reflect a focus on election-related news which begins in early June and continues through the elections and then an introduction of new topics after their end. Again examining the topic distributions, we see that for Oushinet the period before and during the election is marked by high pseudo-probabilities for two topics directly related to the parliamentary elections, one topic surrounding the Spanish prime minister, and two on Russia and Ukraine and the Israel-Palestine war (Appendix \ref{sec:topic-distributions}, Figure \ref{fig:parliament-topics}). Interestingly, pseudo-probabilities for the topic most directly focused on the elections continued to grow even after the election, suggesting that Oushinet was still discussing the election results during this time. Similarly, for Xinouzhou, three topics focused on the UK elections, Europe broadly, and the Spanish prime minister were comparatively prominent towards the end of May and beginning of June.

Overall, we find that this pipeline allows us to effectively locate changes in news ecosystems, correlate these changes to political and cultural events of interest, and further explore possible reasons for these changes via topic models. It reveals differences in media responses both between events and between sites, while also demonstrating the similarities in sites' news ecosystems, such as the increased discussion of the Spanish prime minister on both Xinouzhou and Oushinet before the EU parliamentary elections. We believe that the combination of the novelty and resonance metrics with the novel KeyNMF topic model will permit further in-depth analysis of these media sites and facilitate research on other Chinese-language domains.

\section{Conclusion}

In this paper, we present a pipeline designed to facilitate research on the underlying information dynamics of Chinese diaspora media published in Europe. This pipeline combines existing information-theoretic methods that model how new information enters and persists in systems with a novel topic model, KeyNMF. KeyNMF overcomes some of the weaknesses of previous traditional and contextual topic models, demonstrating high performance on standard benchmarks. We validate this pipeline through preliminary experimentation on our dataset of Chinese diaspora media, finding that it reveals informational trends that correlate with major, newsworthy events in European politics and allows for further analysis of the topical changes that cause those trends. While further qualitative research is required to fully understand these dynamics, we believe that we have presented a major step forward in terms of context-sensitive and interpretable topic modelling and information dynamics which can generalize to multilingual and data scarce environments.

\begin{acknowledgments}
Part of the computation done for this project was performed on the UCloud interactive HPC system, which is managed by the eScience Center at the University of Southern Denmark.
\end{acknowledgments}

\bibliography{bibliography}

\begin{thebibliography}{32}
\expandafter\ifx\csname natexlab\endcsname\relax\def\natexlab#1{#1}\fi
\providecommand{\url}[1]{\texttt{#1}}
\providecommand{\href}[2]{#2}
\providecommand{\path}[1]{#1}
\providecommand{\DOIprefix}{doi:}
\providecommand{\ArXivprefix}{arXiv:}
\providecommand{\URLprefix}{URL: }
\providecommand{\Pubmedprefix}{pmid:}
\providecommand{\doi}[1]{\href{http://dx.doi.org/#1}{\path{#1}}}
\providecommand{\Pubmed}[1]{\href{pmid:#1}{\path{#1}}}
\providecommand{\bibinfo}[2]{#2}
\ifx\xfnm\relax \def\xfnm[#1]{\unskip,\space#1}\fi
\bibitem[{Gatterman et~al.(2022)Gatterman, Meyer, and Wurzer}]{gatterman_2022}
\bibinfo{author}{K.~Gatterman}, \bibinfo{author}{T.~M. Meyer}, \bibinfo{author}{K.~Wurzer},
\newblock \bibinfo{title}{{Who Won the Election? Explaining News Coverage of Election Results in Multi-Party Systems}},
\newblock \bibinfo{journal}{European Journal of Political Research} \bibinfo{volume}{61} (\bibinfo{year}{2022}) \bibinfo{pages}{857--877}. \DOIprefix\doi{10.1111/1475-6765.12498}.
\bibitem[{Schliebs et~al.(2021)Schliebs, Bailey, Bright, and Howard}]{schliebs2021}
\bibinfo{author}{M.~Schliebs}, \bibinfo{author}{H.~Bailey}, \bibinfo{author}{J.~Bright}, \bibinfo{author}{P.~N. Howard}, \bibinfo{title}{{China’s Public Diplomacy Operations: Understanding Engagement and Inauthentic Amplification of PRC Diplomats on Facebook and Twitter}}, \bibinfo{type}{Technical Report}, Oxford, UK: Programme on Democracy \& Technology, \bibinfo{year}{2021}.
\bibitem[{Thunø and Nielbo(2024)}]{thuno2024initial}
\bibinfo{author}{M.~Thunø}, \bibinfo{author}{K.~L. Nielbo},
\newblock \bibinfo{title}{{The Initial Digitalization of Chinese Diplomacy (2019–2021): Establishing Global Communication Networks on Twitter}},
\newblock \bibinfo{journal}{Journal of Contemporary China} \bibinfo{volume}{33} (\bibinfo{year}{2024}) \bibinfo{pages}{244--266}. \DOIprefix\doi{10.1080/10670564.2023.2195811}.
\bibitem[{Brussee(2022)}]{brussee_2022}
\bibinfo{author}{V.~Brussee},
\newblock \bibinfo{title}{{Authoritarian Design: How the Digital Architecture on China’s Sina Weibo Facilitate Information Control}},
\newblock \bibinfo{journal}{Asiascape: Digital Asia} \bibinfo{volume}{9} (\bibinfo{year}{2022}) \bibinfo{pages}{207--241}. \DOIprefix\doi{10.1163/22142312-bja10033}.
\bibitem[{Chan and Alden(2023)}]{chan_alden_2023}
\bibinfo{author}{K.~Chan}, \bibinfo{author}{C.~Alden},
\newblock \bibinfo{title}{{<Redirecting> the Diaspora: China's United Front Work and the Hyperlink Networks of Diasporic Chinese Websites in Cyberspace}},
\newblock \bibinfo{journal}{Political Research Exchange} \bibinfo{volume}{5} (\bibinfo{year}{2023}) \bibinfo{pages}{1--21}. \DOIprefix\doi{10.1080/2474736X.2023.2179409}.
\bibitem[{Barron et~al.(2018)Barron, Huang, Spang, and DeDeo}]{barron-2018-individuals}
\bibinfo{author}{A.~T.~J. Barron}, \bibinfo{author}{J.~Huang}, \bibinfo{author}{R.~L. Spang}, \bibinfo{author}{S.~DeDeo},
\newblock \bibinfo{title}{{Individuals, Institutions, and Innovation in the Debates of the French Revolution}},
\newblock \bibinfo{journal}{Proceedings of the National Academy of Sciences} \bibinfo{volume}{115} (\bibinfo{year}{2018}) \bibinfo{pages}{4607--4612}. \DOIprefix\doi{10.1073/pnas.1717729115}.
\bibitem[{Nielbo et~al.(2020)Nielbo, Vahlstrup, Bechmann, and Gao}]{nielbo2020trend}
\bibinfo{author}{K.~L. Nielbo}, \bibinfo{author}{P.~B. Vahlstrup}, \bibinfo{author}{A.~Bechmann}, \bibinfo{author}{J.~Gao},
\newblock \bibinfo{title}{{Trend Reservoir Detection: Minimal Persistence and Resonant Behavior of Trends in Social Media}},
\newblock in: \bibinfo{booktitle}{Proceedings of the Workshop on Computational Humanities Research (CHR 2020)}, \bibinfo{address}{Amsterdam, the Netherlands}, \bibinfo{year}{2020}, pp. \bibinfo{pages}{290--297}. \DOIprefix\doi{10.48550/arXiv.2109.08589}.
\bibitem[{Nielbo et~al.(2021{\natexlab{a}})Nielbo, Baglini, Vahlstrup, Enevoldsen, Bechmann, and Roepstorff}]{nielbo2021news}
\bibinfo{author}{K.~L. Nielbo}, \bibinfo{author}{R.~B. Baglini}, \bibinfo{author}{P.~B. Vahlstrup}, \bibinfo{author}{K.~C. Enevoldsen}, \bibinfo{author}{A.~Bechmann}, \bibinfo{author}{A.~Roepstorff},
\newblock \bibinfo{title}{{News Information Decoupling: An Information Signature of Catastrophes in Legacy News Media}},
\newblock in: \bibinfo{booktitle}{Proceedings of the 2020 European Association for Digital Humanities Conference}, \bibinfo{address}{Krasnoyarsk, Russia}, \bibinfo{year}{2021}{\natexlab{a}}, pp. \bibinfo{pages}{1--8}. \DOIprefix\doi{10.48550/arXiv.2101.02956}.
\bibitem[{Nielbo et~al.(2021{\natexlab{b}})Nielbo, Haestrup, Enevoldsen, Vahlstrup, Baglini, and Roepstorff}]{nielbo2021no}
\bibinfo{author}{K.~L. Nielbo}, \bibinfo{author}{F.~Haestrup}, \bibinfo{author}{K.~C. Enevoldsen}, \bibinfo{author}{P.~B. Vahlstrup}, \bibinfo{author}{R.~B. Baglini}, \bibinfo{author}{A.~Roepstorff}, \bibinfo{title}{{When No News is Bad News -- Detection of Negative Events from News Media Content}}, \bibinfo{year}{2021}{\natexlab{b}}. \DOIprefix\doi{10.48550/arXiv.2102.06505}. \href{http://arxiv.org/abs/2102.06505}{{\tt arXiv:2102.06505}}.
\bibitem[{Nielbo et~al.(2023)Nielbo, Enevoldsen, Baglini, Fano, Roepstorff, and Gao}]{nielbo2023pandemic}
\bibinfo{author}{K.~L. Nielbo}, \bibinfo{author}{K.~Enevoldsen}, \bibinfo{author}{R.~Baglini}, \bibinfo{author}{E.~Fano}, \bibinfo{author}{A.~Roepstorff}, \bibinfo{author}{J.~Gao},
\newblock \bibinfo{title}{{Pandemic News Information Uncertainty —News Dynamics Mirror Differential Response Strategies to COVID-19}},
\newblock \bibinfo{journal}{PLOS ONE} \bibinfo{volume}{18} (\bibinfo{year}{2023}) \bibinfo{pages}{e0278098}. \DOIprefix\doi{10.1371/journal.pone.0278098}.
\bibitem[{Wevers et~al.(2021)Wevers, Kostkan, and Nielbo}]{wevers2021event}
\bibinfo{author}{M.~Wevers}, \bibinfo{author}{J.~Kostkan}, \bibinfo{author}{K.~L. Nielbo},
\newblock \bibinfo{title}{{Event Flow -- How Events Shaped the Flow of the News, 1950-1995}},
\newblock in: \bibinfo{booktitle}{Proceedings of the Third Conference on Computational Humanities Research, CHR 2021}, \bibinfo{address}{Amsterdam, the Netherlands}, \bibinfo{year}{2021}, pp. \bibinfo{pages}{62--76}. \DOIprefix\doi{10.48550/arXiv.2109.08589}.
\bibitem[{Baglini et~al.(2022)Baglini, {\O}stergaard, Larsen, and Nielbo}]{baglini2022emodynamics}
\bibinfo{author}{R.~B. Baglini}, \bibinfo{author}{S.~M. {\O}stergaard}, \bibinfo{author}{S.~N. Larsen}, \bibinfo{author}{K.~L. Nielbo},
\newblock \bibinfo{title}{{Emodynamics: Detecting and Characterizing Pandemic Sentiment Change Points on Danish Twitter}},
\newblock in: \bibinfo{booktitle}{Proceedings of the Fourth Conference on Computational Humanities Research, CHR 2022}, \bibinfo{address}{Antwerp, Belgium}, \bibinfo{year}{2022}, pp. \bibinfo{pages}{162--176}.
\bibitem[{Blei et~al.(2003)Blei, Ng, and Jordan}]{blei_lda}
\bibinfo{author}{D.~M. Blei}, \bibinfo{author}{A.~Y. Ng}, \bibinfo{author}{M.~I. Jordan},
\newblock \bibinfo{title}{{Latent Dirichlet Allocation}},
\newblock \bibinfo{journal}{Journal of Machine Learning Research} \bibinfo{volume}{3} (\bibinfo{year}{2003}) \bibinfo{pages}{993--1022}. \DOIprefix\doi{10.5555/944919.944937}.
\bibitem[{Blei(2012)}]{blei_prob_topic_models}
\bibinfo{author}{D.~M. Blei},
\newblock \bibinfo{title}{{Probabilistic Topic Models}},
\newblock \bibinfo{journal}{Communications of the ACM} \bibinfo{volume}{55} (\bibinfo{year}{2012}) \bibinfo{pages}{77--84}. \DOIprefix\doi{10.1145/2133806.2133826}.
\bibitem[{Kardos et~al.(2024)Kardos, Kostkan, Vermillet, Nielbo, Enevoldsen, and Rocca}]{kardos2024s3semanticsignal}
\bibinfo{author}{M.~Kardos}, \bibinfo{author}{J.~Kostkan}, \bibinfo{author}{A.-Q. Vermillet}, \bibinfo{author}{K.~Nielbo}, \bibinfo{author}{K.~Enevoldsen}, \bibinfo{author}{R.~Rocca}, \bibinfo{title}{{$S^3$ -- Semantic Signal Separation}}, \bibinfo{year}{2024}. \DOIprefix\doi{10.48550/arXiv.2406.09556}. \href{http://arxiv.org/abs/2406.09556}{{\tt arXiv:2406.09556}}.
\bibitem[{Wallach et~al.(2009)Wallach, Mimno, and McCallum}]{Wallach2009RethinkingLW}
\bibinfo{author}{H.~M. Wallach}, \bibinfo{author}{D.~Mimno}, \bibinfo{author}{A.~McCallum},
\newblock \bibinfo{title}{{Rethinking LDA: Why Priors Matter}},
\newblock in: \bibinfo{booktitle}{Advances in Neural Information Processing Systems}, \bibinfo{address}{Vancouver, Canada}, \bibinfo{year}{2009}, pp. \bibinfo{pages}{1--9}.
\bibitem[{{\v R}eh{\r u}{\v r}ek and Sojka(2010)}]{gensim}
\bibinfo{author}{R.~{\v R}eh{\r u}{\v r}ek}, \bibinfo{author}{P.~Sojka},
\newblock \bibinfo{title}{{Software Framework for Topic Modelling with Large Corpora}},
\newblock in: \bibinfo{booktitle}{{Proceedings of the LREC 2010 Workshop on New Challenges for NLP Frameworks}}, \bibinfo{address}{Valletta, Malta}, \bibinfo{year}{2010}, pp. \bibinfo{pages}{45--50}.
\bibitem[{Pedregosa et~al.(2011)Pedregosa, Varoquaux, Gramfort, Michel, Thirion, Grisel, Blondel, Prettenhofer, Weiss, Dubourg, Vanderplas, Passos, Cournapeau, Brucher, Perrot, and Duchesnay}]{scikit-learn}
\bibinfo{author}{F.~Pedregosa}, \bibinfo{author}{G.~Varoquaux}, \bibinfo{author}{A.~Gramfort}, \bibinfo{author}{V.~Michel}, \bibinfo{author}{B.~Thirion}, \bibinfo{author}{O.~Grisel}, \bibinfo{author}{M.~Blondel}, \bibinfo{author}{P.~Prettenhofer}, \bibinfo{author}{R.~Weiss}, \bibinfo{author}{V.~Dubourg}, \bibinfo{author}{J.~Vanderplas}, \bibinfo{author}{A.~Passos}, \bibinfo{author}{D.~Cournapeau}, \bibinfo{author}{M.~Brucher}, \bibinfo{author}{M.~Perrot}, \bibinfo{author}{E.~Duchesnay},
\newblock \bibinfo{title}{{Scikit-Learn: Machine Learning in Python}},
\newblock \bibinfo{journal}{Journal of Machine Learning Research} \bibinfo{volume}{12} (\bibinfo{year}{2011}) \bibinfo{pages}{2825--2830}.
\bibitem[{Zhao et~al.(2011)Zhao, Qin, and Wan}]{zhao_2011}
\bibinfo{author}{Q.~Zhao}, \bibinfo{author}{Z.~Qin}, \bibinfo{author}{T.~Wan},
\newblock \bibinfo{title}{{Topic Modeling of Chinese Language Using Character-Word Relations}},
\newblock in: \bibinfo{booktitle}{Neural Information Processing}, \bibinfo{address}{Berlin, Heidelberg}, \bibinfo{year}{2011}, pp. \bibinfo{pages}{139--147}.
\bibitem[{Qin et~al.(2016)Qin, Cong, and Wan}]{QIN201660}
\bibinfo{author}{Z.~Qin}, \bibinfo{author}{Y.~Cong}, \bibinfo{author}{T.~Wan},
\newblock \bibinfo{title}{Topic modeling of chinese language beyond a bag-of-words},
\newblock \bibinfo{journal}{Computer Speech \& Language} \bibinfo{volume}{40} (\bibinfo{year}{2016}) \bibinfo{pages}{60--78}. \DOIprefix\doi{https://doi.org/10.1016/j.csl.2016.03.004}.
\bibitem[{Blei and Lafferty(2006)}]{blei_dynamic}
\bibinfo{author}{D.~M. Blei}, \bibinfo{author}{J.~D. Lafferty},
\newblock \bibinfo{title}{{Dynamic Topic Models}},
\newblock in: \bibinfo{booktitle}{Proceedings of the 23rd International Conference on Machine Learning}, \bibinfo{address}{Pittsburgh, Pennsylvania, USA}, \bibinfo{year}{2006}, pp. \bibinfo{pages}{113--120}. \DOIprefix\doi{10.1145/1143844.1143859}.
\bibitem[{Reimers and Gurevych(2019)}]{reimers-2019-sentence-bert}
\bibinfo{author}{N.~Reimers}, \bibinfo{author}{I.~Gurevych},
\newblock \bibinfo{title}{{Sentence-BERT: Sentence Embeddings using Siamese BERT-Networks}},
\newblock in: \bibinfo{booktitle}{Proceedings of the 2019 Conference on Empirical Methods in Natural Language Processing}, \bibinfo{address}{Hong Kong, China}, \bibinfo{year}{2019}, pp. \bibinfo{pages}{3982--3992}. \DOIprefix\doi{10.18653/v1/D19-1410}.
\bibitem[{Bianchi et~al.(2021{\natexlab{a}})Bianchi, Terragni, and Hovy}]{ctm}
\bibinfo{author}{F.~Bianchi}, \bibinfo{author}{S.~Terragni}, \bibinfo{author}{D.~Hovy},
\newblock \bibinfo{title}{{Pre-Training is a Hot Topic: Contextualized Document Embeddings Improve Topic Coherence}},
\newblock in: \bibinfo{booktitle}{Proceedings of the 59th Annual Meeting of the Association for Computational Linguistics and the 11th International Joint Conference on Natural Language Processing (Volume 2: Short Papers)}, \bibinfo{address}{Online}, \bibinfo{year}{2021}{\natexlab{a}}, pp. \bibinfo{pages}{759--766}. \DOIprefix\doi{10.18653/v1/2021.acl-short.96}.
\bibitem[{Bianchi et~al.(2021{\natexlab{b}})Bianchi, Terragni, Hovy, Nozza, and Fersini}]{zeroshot_tm}
\bibinfo{author}{F.~Bianchi}, \bibinfo{author}{S.~Terragni}, \bibinfo{author}{D.~Hovy}, \bibinfo{author}{D.~Nozza}, \bibinfo{author}{E.~Fersini},
\newblock \bibinfo{title}{{Cross-lingual Contextualized Topic Models with Zero-shot Learning}},
\newblock in: \bibinfo{booktitle}{Proceedings of the 16th Conference of the European Chapter of the Association for Computational Linguistics: Main Volume}, \bibinfo{address}{Online}, \bibinfo{year}{2021}{\natexlab{b}}, pp. \bibinfo{pages}{1676--1683}. \DOIprefix\doi{10.18653/v1/2021.eacl-main.143}.
\bibitem[{Grootendorst(2022)}]{bertopic_paper}
\bibinfo{author}{M.~Grootendorst}, \bibinfo{title}{{BERTopic: Neural Topic Modeling with a Class-Based TF-IDF Procedure}}, \bibinfo{year}{2022}. \DOIprefix\doi{10.48550/arXiv.2203.05794}. \href{http://arxiv.org/abs/2203.05794}{{\tt arXiv:2203.05794}}.
\bibitem[{Angelov(2020)}]{top2vec}
\bibinfo{author}{D.~Angelov}, \bibinfo{title}{{Top2Vec: Distributed Representations of Topics}}, \bibinfo{year}{2020}. \href{http://arxiv.org/abs/2008.09470}{{\tt arXiv:2008.09470}}.
\bibitem[{Belford et~al.(2018)Belford, Mac~Namee, and Greene}]{nmf_stability}
\bibinfo{author}{M.~Belford}, \bibinfo{author}{B.~Mac~Namee}, \bibinfo{author}{D.~Greene},
\newblock \bibinfo{title}{{Stability of Topic Modeling via Matrix Factorization}},
\newblock \bibinfo{journal}{Expert Systems With Applications} \bibinfo{volume}{91} (\bibinfo{year}{2018}) \bibinfo{pages}{159--169}. \DOIprefix\doi{10.1016/j.eswa.2017.08.047}.
\bibitem[{Lefèvre et~al.(2011)Lefèvre, Bach, and Févotte}]{online_nmf}
\bibinfo{author}{A.~Lefèvre}, \bibinfo{author}{F.~Bach}, \bibinfo{author}{C.~Févotte},
\newblock \bibinfo{title}{{Online Algorithms for Nonnegative Matrix Factorization with the Itakura-Saito Divergence}},
\newblock in: \bibinfo{booktitle}{2011 IEEE Workshop on Applications of Signal Processing to Audio and Acoustics (WASPAA)}, \bibinfo{address}{New Paltz, NY, USA}, \bibinfo{year}{2011}, pp. \bibinfo{pages}{313--316}. \DOIprefix\doi{10.1109/ASPAA.2011.6082314}.
\bibitem[{Cichocki and Phan(2009)}]{nmf_impl}
\bibinfo{author}{A.~Cichocki}, \bibinfo{author}{A.-H. Phan},
\newblock \bibinfo{title}{{Fast Local Algorithms for Large Scale Nonnegative Matrix and Tensor Factorizations}},
\newblock \bibinfo{journal}{IEICE Transactions on Fundamentals of Electronics, Communications and Computer Sciences} \bibinfo{volume}{E92.A} (\bibinfo{year}{2009}) \bibinfo{pages}{708--721}. \DOIprefix\doi{10.1587/transfun.E92.A.708}.
\bibitem[{Grootendorst(2020)}]{grootendorst2020keybert}
\bibinfo{author}{M.~Grootendorst}, \bibinfo{title}{{KeyBERT: Minimal Keyword Extraction with BERT}}, \bibinfo{year}{2020}. \DOIprefix\doi{10.5281/zenodo.4461265}.
\bibitem[{Reimers and Gurevych(2020)}]{reimers-2020-multilingual-sentence-bert}
\bibinfo{author}{N.~Reimers}, \bibinfo{author}{I.~Gurevych},
\newblock \bibinfo{title}{{Making Monolingual Sentence Embeddings Multilingual using Knowledge Distillation}},
\newblock in: \bibinfo{booktitle}{Proceedings of the 2020 Conference on Empirical Methods in Natural Language Processing}, \bibinfo{address}{Online}, \bibinfo{year}{2020}, pp. \bibinfo{pages}{4512--4525}. \DOIprefix\doi{10.48550/arXiv.2004.09813}.
\bibitem[{Pöttker(2003)}]{pottker_2003}
\bibinfo{author}{H.~Pöttker},
\newblock \bibinfo{title}{{News and Its Communicative Quality: the Inverted Pyramid — When and Why Did It Appear?}},
\newblock \bibinfo{journal}{Journalism Studies} \bibinfo{volume}{4} (\bibinfo{year}{2003}) \bibinfo{pages}{501--511}. \DOIprefix\doi{10.1080/1461670032000136596}.

\end{thebibliography}
\pagebreak
\appendix

\section{News Site Subpages}
\label{sec:newsSubpages}

The subpages scraped for each news site are listed below:
\begin{itemize}
    \item \textbf{Xinouzhou:} France, Italy, Spain, UK, Germany, Hungary, International
    \item \textbf{Ihuawen:} News, Comments \& Opinions
    \item \textbf{Oushinet:} Europe, Europe: Germany, Europe: Central and Eastern Europe, Europe: Italy, Europe: Spain, Europe: Other, France, Europe and China, Overseas Chinese community, China, International, Opinion on public affairs
    \item \textbf{Chinanews:} Nordic headlines, China news, Mutual learning among civilizations, Overseas Chinese community, Nordic Commercial Bridge, Overseas thoughts
    \item \textbf{Yidali Huarenjie:} $\varnothing$
\end{itemize}

\section{NPMI Coherence}
Since NPMI Coherence has historical significance in topic modeling literature, we also evaluated topic descriptions with this metrics.
Due to theoretical and practical limitations\citep{kardos2024s3semanticsignal}, however, we do not consider NPMI Coherence a good metric for evaluating topic models.
For the sake of completeness, we report $C_{\text{NPMI}}$ scores in Table \ref{tab:c_npmi}.

\begin{table}[h]
\caption{$C_{\text{NPMI}}$ coherence of different topic models on the studied corpora.}
\begin{tabular}{l|ccccc}
\toprule
\textbf{Model} & chinanews & ihuawen & oushinet & xinozhou & yidali-huarenjie \\
\midrule
BERTopic & \underline{0.06} & \textbf{0.11} & \underline{0.08} & \textbf{0.10} & \underline{0.08} \\
CombinedTM & -0.07 & -0.02 & -0.07 & -0.02 & -0.12 \\
KeyNMF & -0.21 & -0.23 & -0.00 & -0.04 & -0.14 \\
LDA & -0.02 & -0.02 & 0.02 & \underline{0.03} & 0.00 \\
NMF & \textbf{0.11} & \underline{0.08} & \textbf{0.10} & \textbf{0.10} & \textbf{0.10} \\
S³ & -0.37 & -0.37 & -0.22 & -0.18 & -0.37 \\
Top2Vec & -0.36 & -0.36 & -0.25 & -0.22 & -0.36 \\
ZeroShotTM & -0.07 & -0.03 & -0.06 & -0.03 & -0.11 \\
\bottomrule
\end{tabular}
\label{tab:c_npmi}
\end{table}

\pagebreak
\section{Additional Experimental Results}
\subsection{Novelty and Resonance Ablations}
\label{sec:novelty-resonance-ablations}
\begin{figure}[!htbp]
\centering
\includegraphics[scale=0.18]{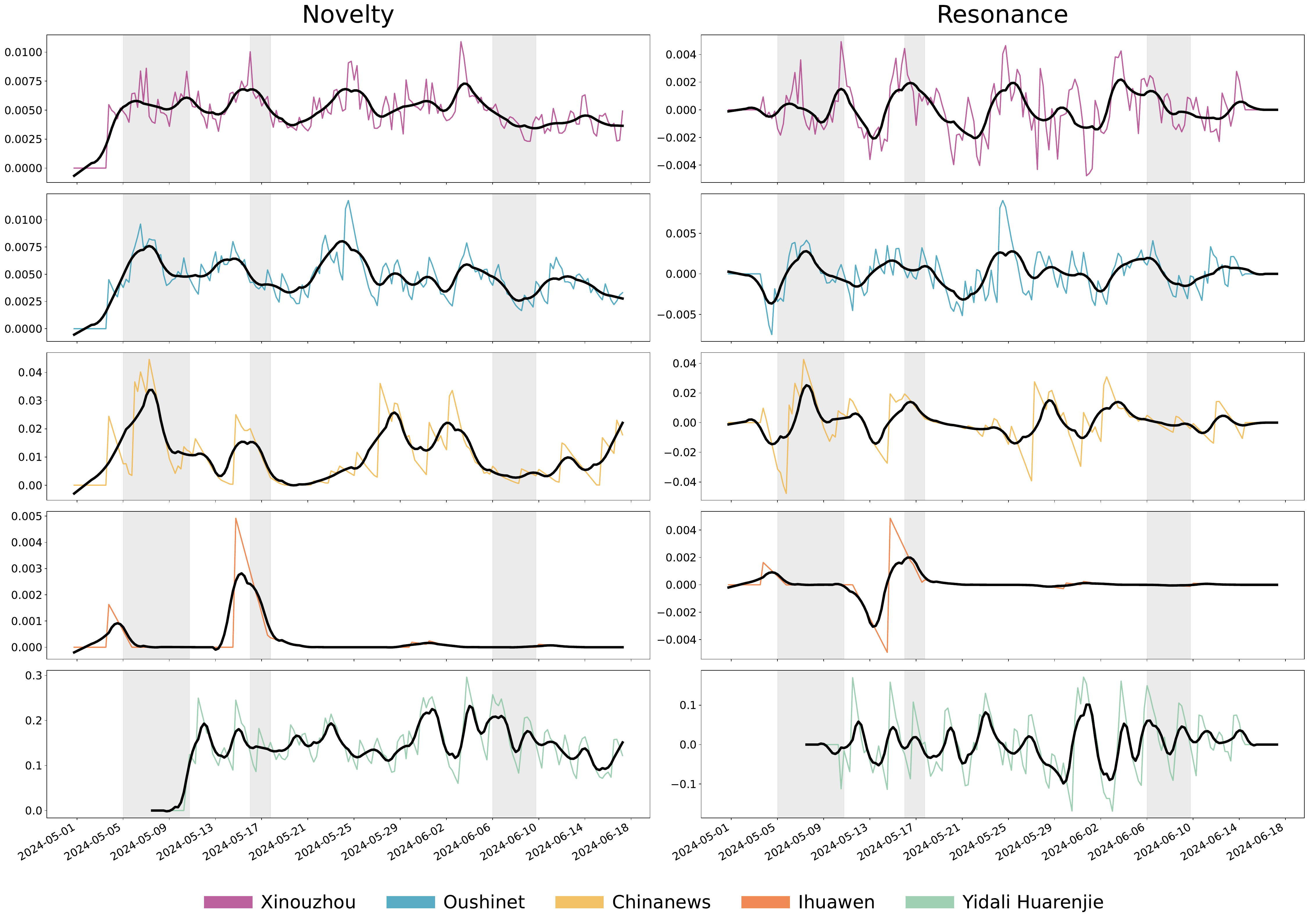}
\label{fig:nov-res-25}
\caption{The novelty and resonance plots for each news site from KeyNMF with 25 topics. The three shaded areas represent Xi Jinping's European tour (May 5-10, 2024), Putin's state visit to China (May 16-17, 2024), and the EU parliamentary elections (June 6-9, 2024). Note that the y-axis ranges differ for each chart.}
\end{figure}

\begin{figure}[!htbp]
\centering
\includegraphics[scale=0.18]{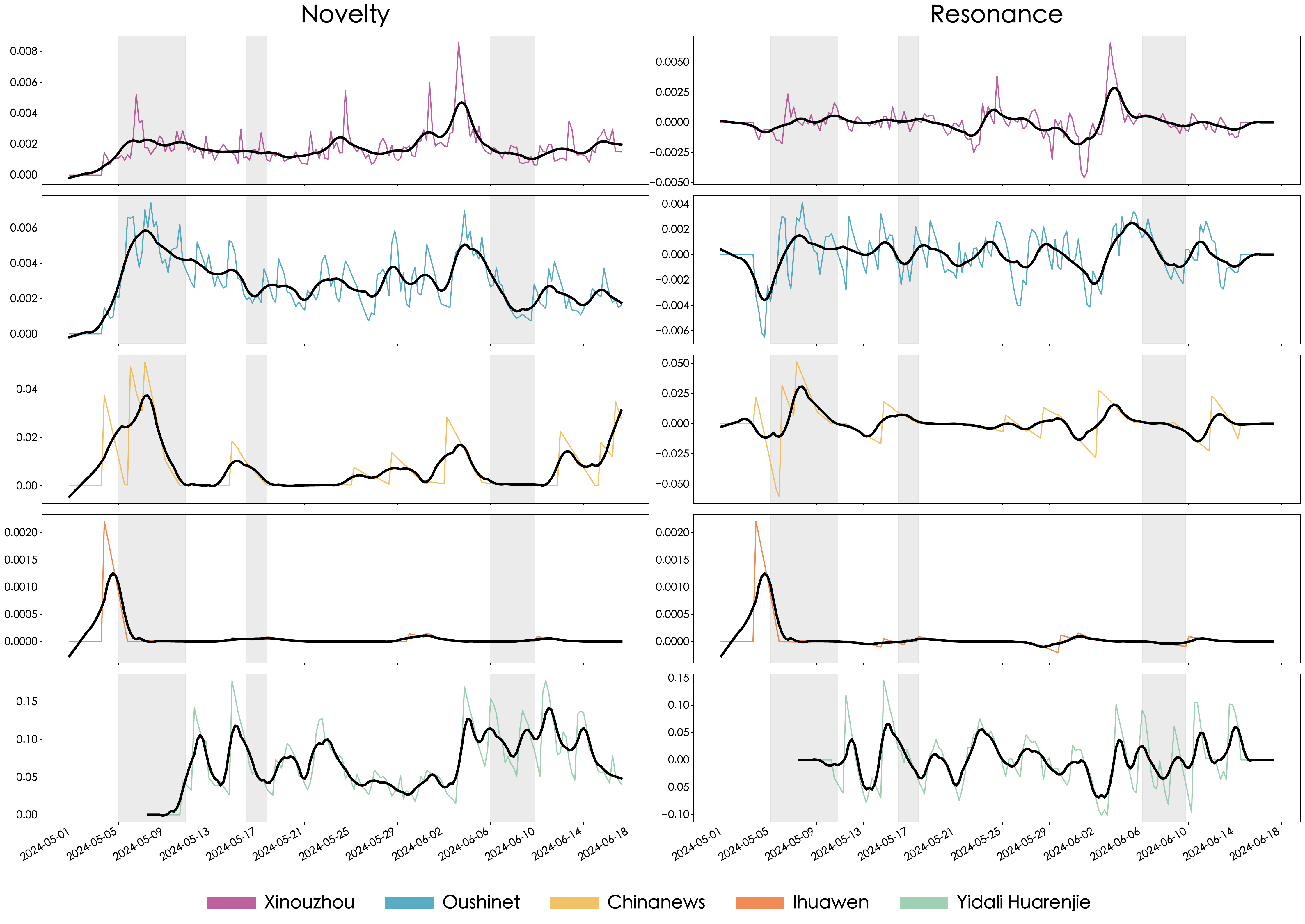}
\label{fig:nov-res-50}
\caption{The novelty and resonance plots for each news site from KeyNMF with 50 topics. The three shaded areas represent Xi Jinping's European tour (May 5-10, 2024), Putin's state visit to China (May 16-17, 2024), and the EU parliamentary elections (June 6-9, 2024). Note that the y-axis ranges differ for each chart.}
\end{figure}

\pagebreak
\subsection{Topic Distributions Over Time}
\label{sec:topic-distributions}
\begin{figure}[!htbp]
\centering
\includegraphics[scale=0.25]{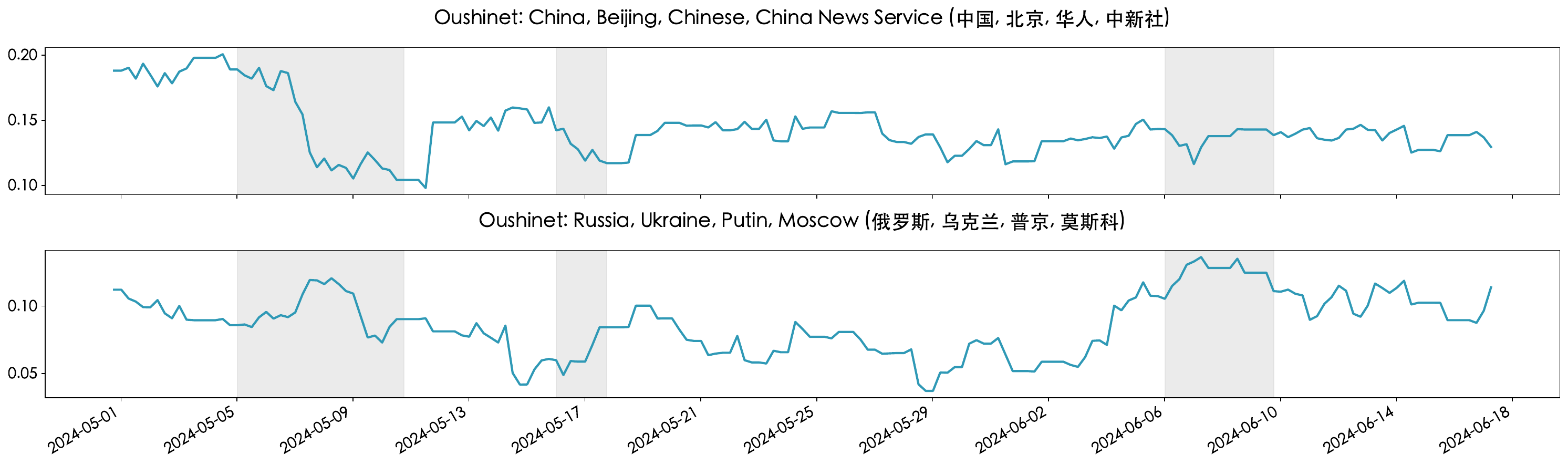}
\caption{The distributions over time for two topics with high pseudo-probabilities before Putin's state visit to China. These topics are generated by the 10-topic KeyNMF model for Oushinet. Note that the y-axis scale differs for each subplot.}
\label{fig:state-visit-topics}
\end{figure}

\begin{figure}[!htbp]
\centering
\includegraphics[scale=0.25]{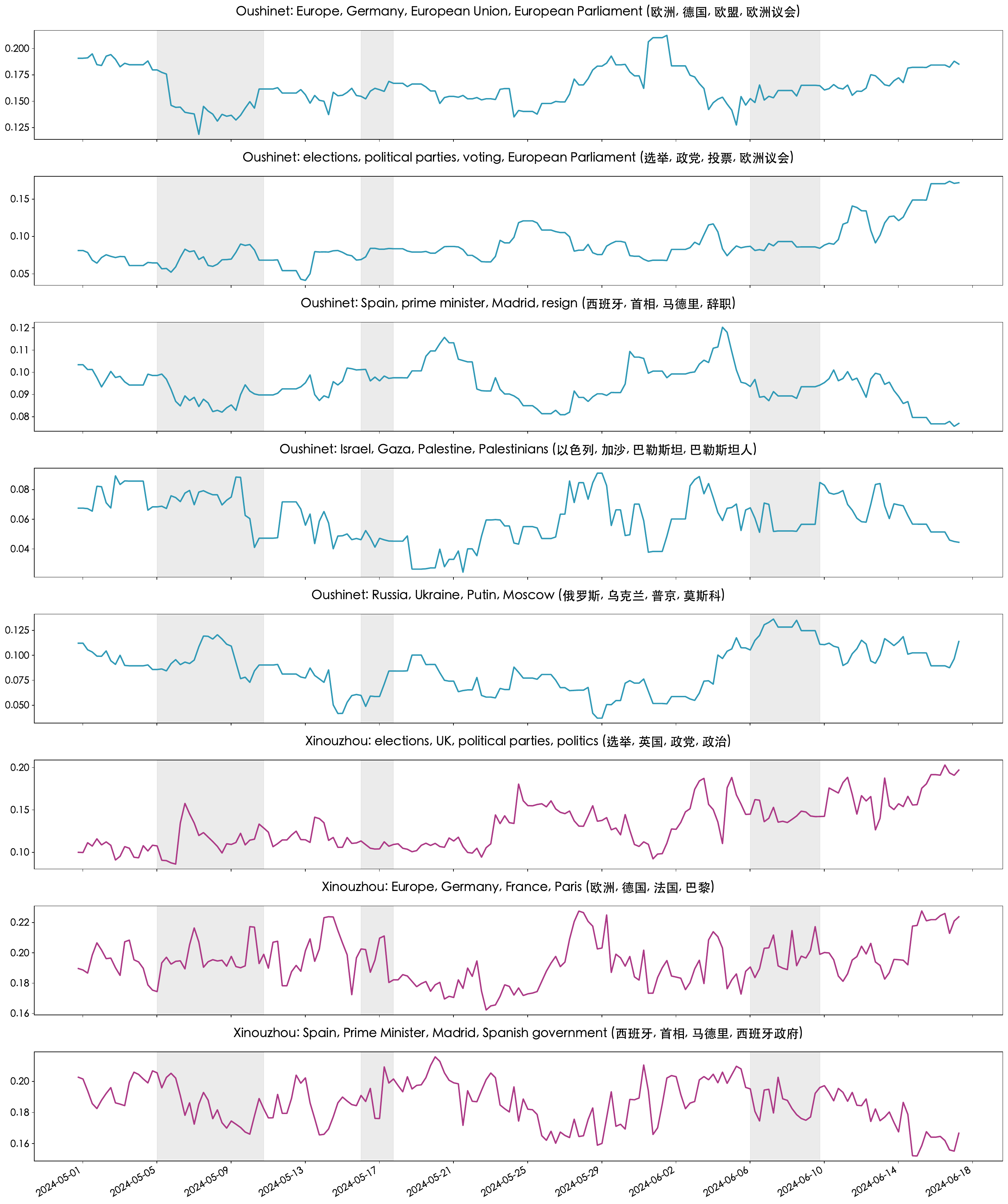}
\caption{The distributions over time for eight topics with high pseudo-probabilities around the EU parliamentary elections. These topics are generated by the 10-topic KeyNMF models for Oushinet and Xinouzhou. Note that the y-axis scale differs for each subplot.}
\label{fig:parliament-topics}
\end{figure}

\begin{figure}[!htbp]
\centering
\includegraphics[scale=0.25]{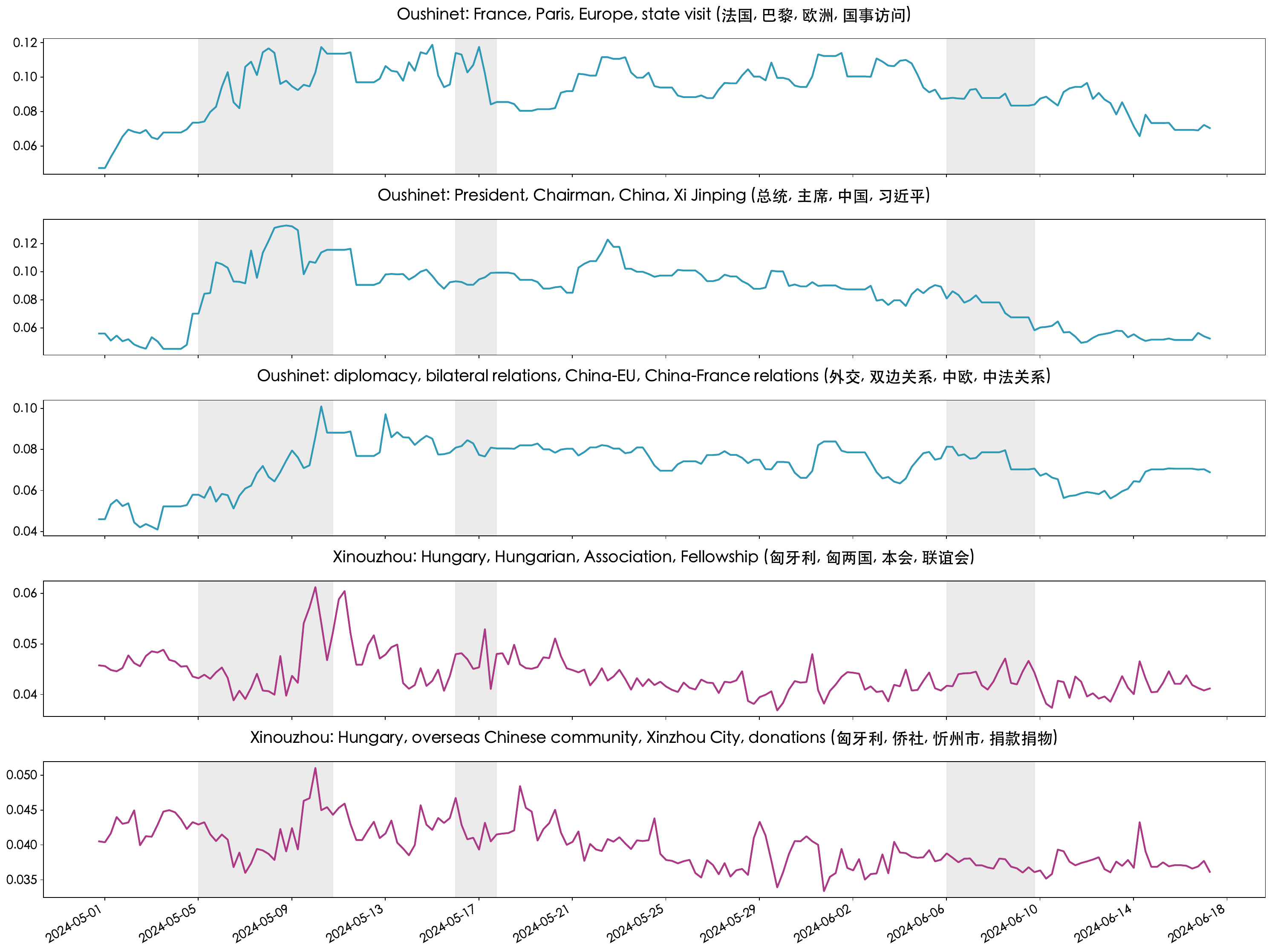}
\caption{The distributions over time for five topics with high pseudo-probabilities during Xi Jinping's European tour. These topics are generated by the 10-topic KeyNMF models for Oushinet and Xinouzhou. Note that the y-axis scale differs for each subplot.}
\label{fig:tour-topics}
\end{figure}

\end{document}